%% file: main.tex
\documentclass[conference]{IEEEtran}
\usepackage{cite}
\usepackage{amsmath,amssymb,amsfonts}
\usepackage{algorithmic}
\usepackage{graphicx}
\usepackage{textcomp}
\usepackage{xcolor}
\usepackage{subfigure}
\usepackage[ruled,vlined]{algorithm2e}
\usepackage{booktabs}
\usepackage{multirow}
\def\BibTeX{{\rm B\kern-.05em{\sc i\kern-.025em b}\kern-.08em
    T\kern-.1667em\lower.7ex\hbox{E}\kern-.125emX}}
\begin{document}

\title{Multi-axis Attentive Prediction for Sparse Event Data:  An Application to Crime Prediction}

\author{\IEEEauthorblockN{Yi Sui}
\IEEEauthorblockA{\textit{yi.sui@mail.utoronto.ca}}
\and
\IEEEauthorblockN{Ga Wu}
\IEEEauthorblockA{\textit{wuga@mie.utoronto.ca}}
\and
\IEEEauthorblockN{Scott Sanner}
\IEEEauthorblockA{\textit{ssanner@mie.utoronto.ca}}}

\maketitle

\begin{abstract}
Spatiotemporal prediction of event data is a challenging task with a long history of research. 
While recent work in spatiotemporal prediction has leveraged deep sequential models that substantially improve over classical approaches, these models are prone to overfitting when the observation is extremely sparse, as in the task of crime event prediction.  
To overcome these sparsity issues, we present Multi-axis Attentive Prediction for Sparse Event Data (MAPSED). We propose a purely attentional approach to extract both short-term dynamics and long-term semantics of event propagation through two observation angles. Unlike existing temporal prediction models that propagate latent information primarily along the temporal dimension, MAPSED simultaneously operates over all axes (time, 2D space, event type) of the embedded data tensor. 
We additionally introduce a novel Frobenius norm-based contrastive learning objective to improve latent representational generalization.
Empirically, we validate MAPSED on two publicly accessible urban crime datasets for spatiotemporal sparse event prediction, where MAPSED outperforms both classical and state-of-the-art deep learning models. The proposed contrastive learning objective significantly enhances  MAPSED's ability to capture the semantics and dynamics of the events, resulting in better generalization ability to combat sparse observations.
\end{abstract}


\section{Introduction}
\input{content/Introduction.tex}
\label{introduction}

\section{Preliminary}
\input{content/Preliminaries.tex}
\label{Preliminary}

\section{MAPSED for Spatiotemporal Event Prediction}
\input{content/Model.tex}

\label{Model}

\section{Experiment and Evaluation}
\input{content/Experiment.tex}
\label{Experiment}

\section{Conclusion}
\input{content/Conclusion}
\label{Conclusion}

\bibliographystyle{IEEEtran}
\bibliography{reference}
\input{appendix}

\end{document}

%% file: content/Introduction.tex
Spatiotemporal Event Prediction is a critical task that involves numerous real-world applications, including precipitation forecasting, urban flow prediction, air quality prediction, crime forecasting, and traffic predictions \cite{convlstm,predcnn,flow,STResNet,crime,TCP,airquality}. The challenging requirements of predicting event occurrence~(in both location and time) prohibit the use of many classical machine learning methods that do not easily model mixed contextual factors often present in this data~\cite{okawa2019deep}.

Previous research has investigated various approaches to support spatiotemporal predictions. Point process-based approaches~\cite{STpointprocess,pointprocess} that leverage probability theory model the sequence of events in continuous time and space.
Alternatively, Gaussian processes~\cite{GP} provide a Bayesian perspective that yields strong performance on small-to-moderate size datasets when kernels are carefully selected and tuned.
Unfortunately, both of these approaches have limitations for real-world data. Point processes make restrictive Poisson assumptions (i.e., mean and variance are equal) on the event occurrence probability, whereas GPs can be computationally prohibitive for large datasets and challenging to tune across the entire range of data due to their intrinsic homoskedastic assumptions.

Recent research ~\cite{convlstm,convlstm-traffic,bradbury2016quasi} introduced modern deep sequential frameworks to model complex event dynamics and achieved strong performance. 
By taking advantage of the flexibility enabled by modern deep learning, these models can learn complex contextual dependencies between event categories, space, and time.
For example, \cite{convlstm} proposes a Convolutional LSTM (ConvLSTM) that combines CNN and LSTM models, which achieved state-of-the-art performance in precipitation nowcasting. 
\cite{convlstm-traffic} introduced similar approaches for traffic accidents prediction.
And, Quasi-RNN~\cite{bradbury2016quasi} interleaved convolutional layers with simple recurrent layers.
However, the drawback of the flexibility of these models is that they may be subject to overfit, as we will demonstrate experimentally.

In this paper, we propose a novel spatiotemporal deep-learning model aiming to achieve strong generalization in the challenging setting of sparse observations. Specifically, we take a tensor-centric, fully attentional approach to extract the short-term dynamics and long-term semantics of event propagation. Concretely, we apply concatenations of the input from two different angles and then use a Multi-axis Attention Block (MAB) to learn the semantics and dynamics. In particular, we conduct depth-wise and breadth-wise concatenation to facilitate the information capturing.
The proposed MAB provides a unified method to capture spatial, temporal, and categorical dependencies. Additionally, we propose a
Frobenius norm based contrastive estimation to facilitate long-term semantics learning.

In our extended experiments, we validate the proposed MAPSED model on two publicly accessible urban crime datasets for spatiotemporal sparse event prediction. We note the MAPSED outperforms both classical and state-of-the-art deep learning models. Remarkably, the MAPSED can combat overfitting problems that commonly happen in the existing models~\cite{crime-transfer}, which shows the significant advantage of MAPSED in transferring the learned crime prediction model to other regions or territories.

%% file: content/Preliminaries.tex
\subsection{Notation}
Here, we list the notation used in the paper.
\begin{itemize}
    \item $c$: Total number of event types.
    \item $h,w$: Total number of geospatial grids along the vertical and horizontal direction.
    \item Historical $m$ event occurrences $\mathbf{X}=[X_1, X_2 ... X_m]$ as input to the model.
    \item Future $n$ event occurrences $\mathbf{Y}=[Y_1, Y_2,...Y_{n}]$ as the ground truth prediction target.
    \item $D$: depth-wise concatenation of $\mathbf{X}$
    \item $S$: breath-wise concatenation of  $\mathbf{X}$
    \item $D', S'$: Extracted dynamics and semantics of $\mathbf{X}$.
    \item $S'^+$: Extracted semantics from the reordered $\mathbf{X}$.
    \item $\mathbf{S'^-}$: Extracted semantics of other sequences in the dataset.
    \item $\mathbf{X}_D$, $\mathbf{X}_S$: $D', S'$ after reverse concatenation.
    \item $\mathbf{U}$: Output of the encoder.
    \item $\mathbf{Y}'$: The predicted future $n$ event occurrences.
    \item $Z, Z'$: Input and output of MAB.
    \item In spatial-axes attention of MAB:
        \begin{itemize}
            \item $Q,K,V$: Attention parameters.
            \item $A=\{a_{i,j}\}$: Attention weights.
            \item $\hat{Z}$: Attended output.
        \end{itemize}
    \item In channel-axis attention of MAB:
        \begin{itemize}
            \item $\tilde{Q},\tilde{K},\tilde{V}$: Attention parameters.
            \item $\tilde{A}=\{\tilde{a}_{i,j}\}$: Attention weights.
            \item $\tilde{Z}$: Attended output.
        \end{itemize}    
    \item $f_d, f_d ^{-1}$: Depth concatenation operation and its inverse
    \item $f_s, f_s^{-1}$: Breath concatenation operation and its inverse
    \item $\Psi_{bn}^{2d}$: 2D convolutional bottleneck layer 
    \item $\Psi_{bn}^{3d}$: 3D convolutional bottleneck layer 
    \item $L,L_r,L_c$: Net loss, reconstruction loss, and contrastive loss
\end{itemize}

\subsection{Spatiotemporal event predictions}
Forecasting  multi-step future events based on the past observations, or Spatiotemporal Event Prediction~(SEP), is a
challenging but attractive task that involves various real-life applications, such as  crime prediction, urban flow prediction, air quality prediction, traffic predictions, and precipitation forecasting \cite{csan,crime,deepcrime,flow,STResNet,airquality,convlstm-traffic,convlstm,predcnn}.

As the observations~(data) are commonly series of events that occur over a period of time, sequential modelling methods, such as ARIMA and LSTM~\cite{ARIMA-crime,lstm-time-series} are preferable to model such occurrences. 
Unfortunately, traditional sequential modelling methods lack the ability to capture spatial information. Thus, much research has focused on exploring the possibility of incorporating spatial modelling methods, such as Convolutional Neural Networks~(CNNs).
Concretely, \cite{convlstm} proposes Convolutional LSTM (ConvLSTM) that combines CNN and LSTM models, which achieved state-of-the-art performance in precipitation nowcasting. Similarly, ~\cite{convGRU} proposes convGRU that combines CNN and GRU models.
Later works further applied the same architecture to closely related domains,
such as demand services prediction \cite{convlstm-on-demand}, crowd flow prediction \cite{convlstm-crowdflow}, and traffic accidents prediction \cite{convlstm-traffic}, where it significantly improved prediction performance.
Besides, ~\cite{Mist} proposes to use LSTM to encode the temporal correlations and then use an attention mechanism for spatial information fusion. 
Similarly, another model, Temporal Convolutional Network, has also been applied to spatiotemporal event predictions \cite{TCN,duronet} as an replacement of RNNs.
In addition, recent research shows the correlations among different event types are also informative for SEP.
XIn particular, CSAN \cite{csan} introduced a CNN layer to fuse different crime types before feeding into the GRU layer. Similarly, Deepcrime \cite{deepcrime} presented a categorical embedding layer before the sequential modules. Also, ~\cite{Cross-Interaction} considers time-category and region-category units after the sequential information encoding.

\subsection{Application of attentive information extraction}
By dynamically adjusting weights based on input observations, the attention mechanism allows a deep model to capture the most informative content for further processing.
This property is beneficial in the sequential prediction tasks~\cite{transformer,bert}, where capturing long-term dependency is challenging, even for the state-of-the-art recurrent architecture, such as LSTM and GRU~\cite{speech}.

Formally, an attention layer extracts predictive abstractions $\mathbf{z}$ from feature map $V=[\mathbf{v}_1,\mathbf{v}_2\cdots\mathbf{v}_n]$ through a weighted linear combination
\begin{equation}
\mathbf{z} = \sum_j^n a_j \mathbf{v}_j,
\end{equation}
where the weights $\mathbf{a}$ are dynamically adjusted as a function of input values $V$. In particular, one of the most widely used weighting functions, Scaled Dot-Product Attention~\cite{transformer}, calculates the attention weights as follows
\begin{equation}
    a_j = \frac{k(\mathbf{v}_i)^T\mathbf{q}_j/\sqrt{d}}{\sum_i^nk(\mathbf{v}_i)^T\mathbf{q}_j/\sqrt{d}},
    \label{eq:scaled-dot-product}
\end{equation}
where $d$ refers to the dimension of vectors $k(\mathbf{v}_i)$ and $\mathbf{q}_j$. To clarify, we note the key vector $k(\mathbf{v}_i)$ is a linear function of input feature $\mathbf{v}_i$.

Aside of the application of Natural Language Processing~(NLP)~\cite{transformer,bert,GPT,speech,sentiment}, attentive information extraction~\cite{transformer} has spread out multiple application domains, including Computer Vision(CV)~\cite{G3AN}, and Recommender Systems(RS) \cite{RS,RS1,attn-survey}. 

Recently, attention mechanism has been frequently applied to problems with sparse observations as it effectively learns from the key information from the sparse observations\cite{RS-attention-sparse, RS-attention-sparse-1, time-series-sparse-attention}. 

\subsection{Contrastive estimation for representation learning}
Contrastive estimation~\cite{contrastive} learns meaningful latent representations by measuring similarity and difference between pairs of samples. 
Specifically, contrastive estimation encourages the latent representations of observations with similar semantic meaning closer to each other than that of observations from different populations.
This method is particularly valuable in the field of self-supervised representation learning research, where data does not have labels.
Remarkably, \cite{simclr,moco} learns a latent representation through contrastive estimation achieved comparable results with supervised learning on image classification. \cite{CPC} adopts contrastive estimation in sequential prediction context that encourages the latent representation to learn long-term features. Contrastive estimation has also been applied to NLP tasks to capture sentence-level semantics and learn a useful representation with unlabelled data\cite{nlp-contrast,cert}.

In the literature, a contrastive estimation objective is commonly expressed as follows: 
\begin{equation}
\begin{aligned}
    &\mathcal{L}_{c} (\mathbf{x}, \mathbf{x^+}, X^-) \\
    &= -\log \frac{\exp(f_k(\psi(\mathbf{x}), \psi(\mathbf{x^+})))}{\exp(f_k(\psi(\mathbf{x}), \psi(\mathbf{x^+})) + \sum_j f_k(\psi(\mathbf{x}),\psi(\mathbf{x}_j^-)))},
\end{aligned}
\end{equation}
where $\mathbf{x}$ denotes a data sample, and $\mathbf{x}^+$ and $X^-$ denote samples that semantically similar and different to the given data sample $\mathbf{x}$ respectively. We use $\psi$ to represent representation learning model and use $f_k$ to represent kernel function. A common choice of the kernel function is simple inner product.

\subsection{Motivation}
By summarizing the existing works in the literature, we note that existing SSF models are usually complex and easy to overfit the training datasets with sparse observations. Therefore, it is challenging to apply those approaches to the crime prediction tasks since the reported crime events in the urban area is exceptionally sparse. To overcome this challenge, we are interested in proposing a novel crime prediction model mitigating the overfitting issue.

In the existing literature, we note that attention models and contrastive learning are effective methods to work with sparse observations~(see \cite{RS-attention-sparse,RS-attention-sparse-1,time-series-sparse-attention,liza2018improving})~\footnote{Attention and contrastive learning are also widely used in the recommendation tasks (see \cite{kang2018self,wu2019noise}), where observations are also sparse.}, which inspired us to explore the possibility of incorporating these techniques into the crime prediction problems. However, during our exploration, we realized the simple combination of existing methods or architectures does not address this particular task. Hence, in this paper, we proposed various modifications to the existing models to facilitate our primary goal -- crime prediction. 

%% file: content/Model.tex
Given a sequence of $m$ consecutive historical event occurrences $\mathbf{X}=[X_1, X_2 ... X_m]$, {\bf Spatiotemporal Event Prediction~(SEP)} aims to predict the event occurrences $\mathbf{Y}=[Y_1, Y_2,...Y_{n}]$ for the next $n$ discretized time steps. Here, an event occurrence record $X_\tau \in \mathbb{N}^{c\times h\times w}$ at time step $\tau$ is a sparse tensor that depicts frequency of events by recording event type $c$ and geographical coordinates $(h,w)$. For a crime prediction task, the event type could be robbery, assault, etc.
Each time step refers to a fixed interval, e.g., a week. 

Note that whereas the classification problem of predicting whether an event would happen is also important ~\cite{Mist,deepcrime,Cross-Interaction}, we focus on the regression task of predicting the event counts. 

\begin{figure}[htbp]
    \centering
    \includegraphics[width=0.95\linewidth]{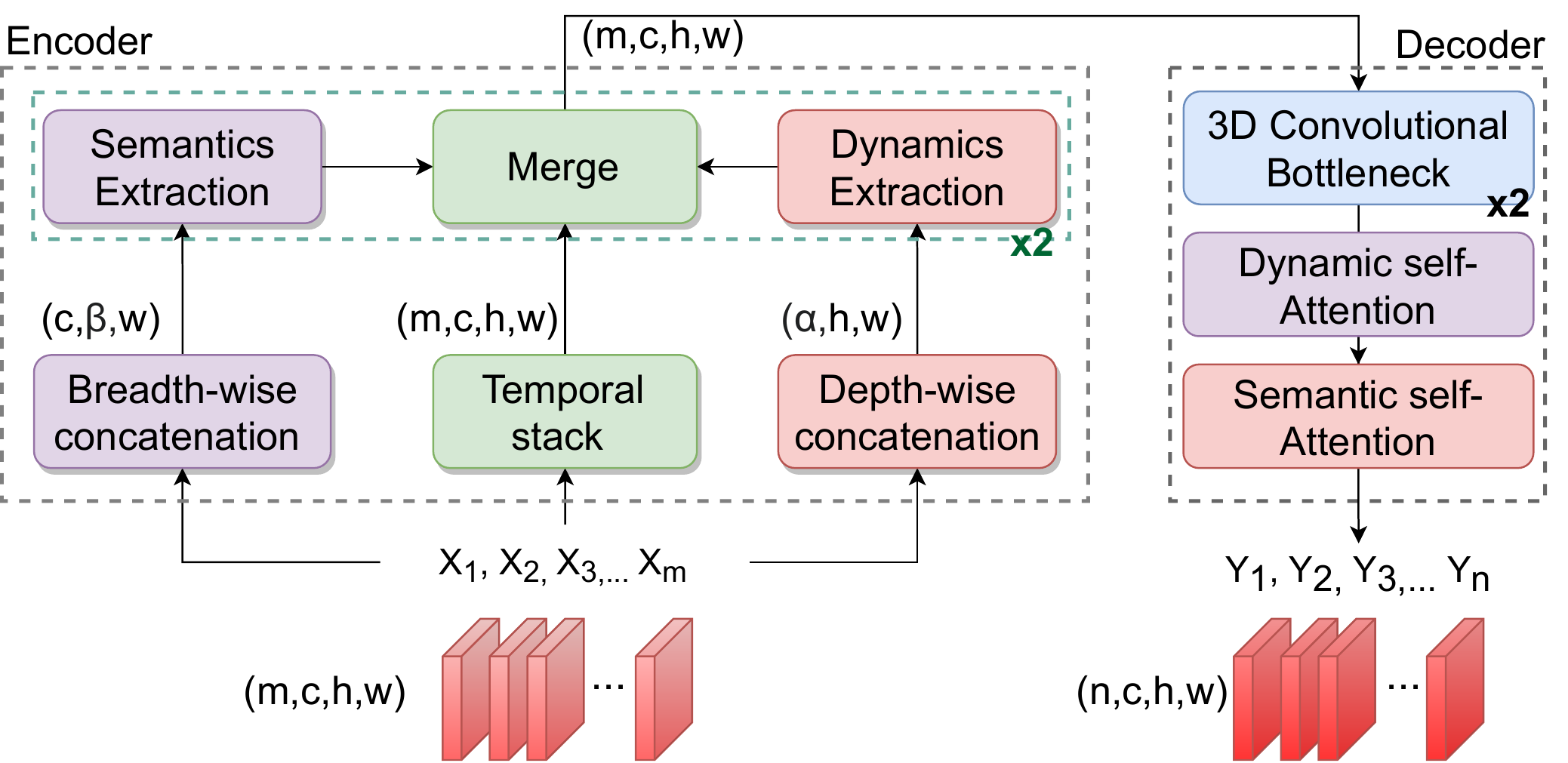}
    \caption{The overall architecture of MAPSED.}
    \label{fig:overall}
\end{figure}

In our MAPSED approach, we choose to generalize the Encoder-Decoder framework for the SEP task.\footnote{In the literature, there are multiple well-established sequential prediction settings. E.g., Tagging Model, Encoder-Decoder, Encoding-Forecasting, etc.} The encoder aims to capture the spatial and temporal information from raw observations $\mathbf{X}$, and the decoder predicts future events $\mathbf{Y}$ by processing captured information. Figure \ref{fig:overall} illustrates the overall architecture of the proposed model. 

We organize the remainder of this section as follows: Section \ref{sec:encoder} and \ref{sec:decoder} detail the encoder and decoder, respectively. Then, we introduce the training objective in \ref{sec:training}.

\subsection{Encoder as an Attentive Information Extraction Model}
\label{sec:encoder}
A sequence of historical event occurrences $\mathbf{X}$ typically carries out two factors of information, namely 1) the short-term dynamics of event development and 2) the time-invariant semantics of correlation among events. Thus, the objective of the encoder is to capture both pieces of information. While it may appear straightforward, effective extraction of the distinct concepts is challenging.

To motivate our approach, a recent study in \emph{recognition science}~\cite{foti2018young} shows that children learn new concepts from the same object by changing their observation angles. We believe the same logic holds for SEP. Thus, we propose to augment the data to mimic the observation angle changes. Specifically, we concatenate the raw observations from two distinct dimensions. 
\begin{itemize}
    \item For capturing dynamics, we conduct depth-wise concatenation, such that the raw observation $\mathbf{X}$ reduces to a 3D tensor $D\in \mathbb{N}^{\alpha\times h\times w}$, where $\alpha = c\times m$. By doing this, we can identify the evolution of events from the channel observation.
    \item Similarly, to capture the time-invariant semantic information, we concatenate raw observation breadth-wise, such that the raw observation $\mathbf{X}$ reduces to a 3D tensor $S\in \mathbb{N}^{c\times \beta\times w}$, where $\beta = m\times h$.
\end{itemize}  

While helpful, the concatenation operations mentioned above do not suffice to encode the raw observations into high-level representations. To efficiently extract information from the concatenated observations $D$ and $S$, we propose a unified network component, namely a Multi-axis Attention Block.

\subsubsection{Multi-axis Attention Block}
\begin{figure}[htbp]
    \centering
    \includegraphics[width=0.98\linewidth]{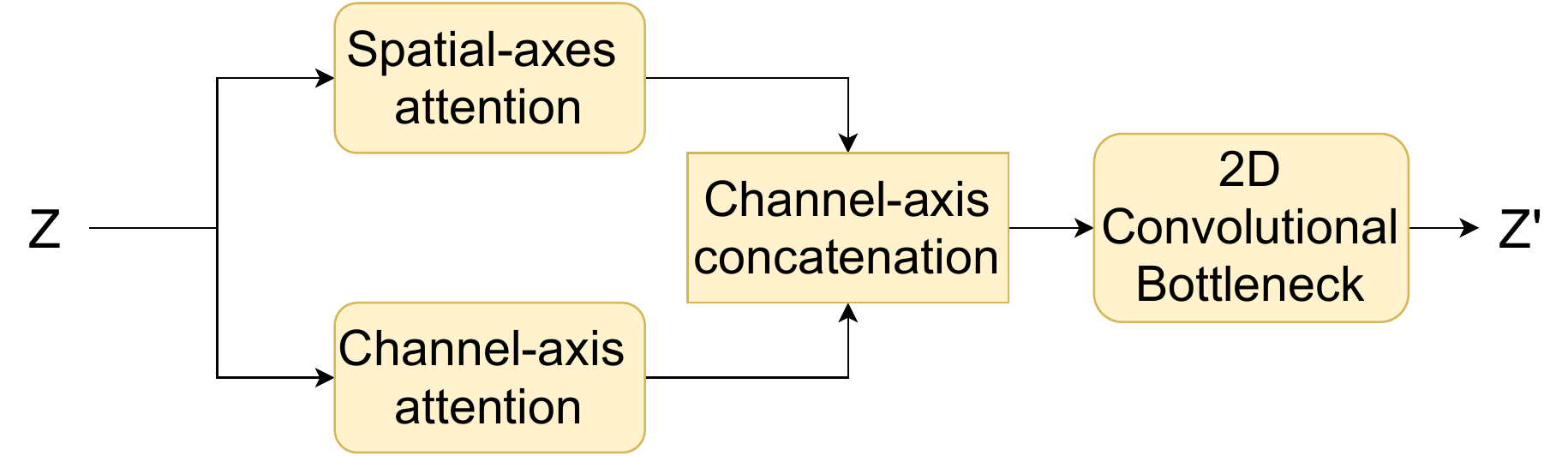}
    \caption{Architecture of the Multi-axis Attention Block(MAB)}
    \label{fig:MAB}
\end{figure}

A Multi-axis Attention Block~(MAB) has three main components: a spatial-axes attention block, a channel-axis attention block, and a fusion function. Figure \ref{fig:MAB} shows the high-level layout of the MAB block.

\noindent{\textit{Spatial Attention}:}
Given the input tensor $Z \in \mathbb{R}^{\tilde{c} \times \tilde{h} \times \tilde{w}}$, spatial attention examines the correlation of events in different geo-locations. To this end, we design the following attention operations:
\begin{align}
\begin{split}
    Q &= \phi_q(Z) \quad K = \phi_k(Z) \quad V = \phi_v(Z)\\
    a_{{i,j}} & = \frac{\exp(\mathbf{q}_{:,i}^T \mathbf{k}_{:,j})}{\sum_{j=1}^{u} \exp(\mathbf{q}_{:,i}^T \mathbf{k}_{:,j})} \\
    \hat{Z} & = V\odot A\\
\end{split}
\end{align}
where $\phi_q, \phi_k, \phi_v$ are convolutional layers with $1\times1$ kernel. We use a single index $i$ or $j$ to represent geo-spacial index in the 2D grids $\tilde{h}\times \tilde{w}$.

Spatial attention described above follows the self-attention definition~\cite{attention}. The difference here is that we aim for 2D attention instead of 1D in the Natural Language Processing literature.

\noindent{\textit{Channel-axis Attention}:}
The channel-axis attention examines the same tensor $Z$ mentioned above from a different angle. Specifically, channel-axis attention defines an alternative attention operation as
\begin{align}
\begin{split}
    \tilde{Q}&=\tilde{K}= \tilde{V} = \phi_c(Z)\\
    \tilde{a}_{i,j} & = \frac{\exp(\tilde{\mathbf{q}}_{i}^T \tilde{\mathbf{k}}_{j})}{\sum_{j=1}^{\tilde{c}} \exp(\tilde{\mathbf{q}}_{i}^T \tilde{\mathbf{k}}_{j})} \\
    \tilde{Z} & = \tilde{A}\tilde{V},\\
\end{split}
\end{align}
where the Query $\tilde{Q}$, Key $\tilde{K}$ and Value $\tilde{V}$ share same values~\cite{DA}. In addition, the attention now examines the correlation of event types. E.g., it estimates how the frequency of theft would influence the chance of robbery happening in the same neighborhood. Note that now we use index $i$ and $j$ to represent the different channels~(event types) instead of grid coordinates. 

\noindent{\textit{Fusion Function}:}
After the two parallel attention operations described above, a fusion function blends the results. Specifically, we stack the tensors ($\hat{Z}$ and $\tilde{Z}$) along the channel axis and pass through a convolutional bottleneck layer
\begin{equation}
    Z' = \Psi_{bn}^{2d}([\hat{Z} || \tilde{Z}])+Z,
    \label{attn_output}
\end{equation}
where $\Psi_{bn}^{2d}$ is a convolutional bottleneck layer composed of 3 sequential convolutional layers with kernel size of  $(1\times 1),(3\times 3), \mathrm{and} \space (1\times 1)$. We use $[\cdot||\cdot]$ to represent channel-wise concatenation. To clarify, we maintain the output $Z'$ as a 3D tensor with same shape of the input $Z$.

\subsubsection{Merging Semantic and Dynamic Information}

With the Multi-axis Attention Block, we process the two concatenated tensors $D$ and $S$ through the same process
\begin{equation}
    D' = \textit{MAB}(D) \quad \textit{and} \quad S' = \textit{MAB}(S),
\label{eq:MAB_encoding}
\end{equation}
where both $D'$ and $S'$ maintain their input's shape. This allows the processed information to reverse the previous concatenation operation and match the shape of raw input $\mathbf{X}$. We denote the reshaped tensor $D'$ and $S'$ as $\mathbf{X}_D$ and $\mathbf{X}_S$. Table~\ref{table:cd_general} summarizes the correlations examined by the MAB-based encoding architecture.

As the last step of the encoding procedure, we merge the information into one tensor $\mathbf{U}$ as the latent representation of observation $\mathbf{X}$. Concretely, the latent representation is a deterministic function
\begin{align}
    \begin{split}
        \mathbf{U} = \Psi_{bn}^{3d}([\mathbf{X}_D||\mathbf{X}_S]) + \Psi_{bn}^{3d}(\mathbf{X}),
    \end{split}
\end{align}
where $\Psi_{bn}^{3d}$ is a 3D convolutional bottleneck layer with kernel size of  $(1\times 1\times 1),(3\times 3\times 3), \textrm{and } \space (1\times 1\times 1)$.

We can further stack multiple layers of the above processing to obtain better performance. In our experiments, we stack two layers.
\begin{table}[t!]
\caption{Correlations examined by the MAB-based encoding architecture}
\resizebox{0.48\textwidth}{!}{%
\begin{tabular}{cccc}
\toprule
Concatenation& Attention Axis & Perception Field & Correlation\\
\midrule
\multirow{2}{*}{Dynamics}&Space&$h\times w$ & Spatial\\\cmidrule(l){2-4} &
Channel&$m\times c$ & Temporal, Categorical\\
\midrule
\multirow{2}{*}{Semantics}&Space&$m\times h\times w$ & Spatiotemporal\\\cmidrule(l){2-4} &
Channel&$c$ & Categorical\\
\bottomrule
\end{tabular}}
\label{table:cd_general}
\end{table}

\subsection{Decoder as parallel predictions}
\label{sec:decoder}

To predict the future events for $n$ time steps, we need to process the encoded latent representation tensor $\mathbf{U} \in \mathbb{R}^{m\times c\times h \times w}$ into a tensor $\hat{\mathbf{Y}} \in \mathbb{R}^{n\times c\times h \times w}$ to match the shape of the ground truth observation $\mathbf{Y}$. Here, we can simply run two stacked 3D convolutional bottleneck layers $\Psi_{bn}^{3d}$ as
\[\hat{\mathbf{Y}} = \Psi_{bn}^{3d}(\Psi_{bn}^{3d}(\mathbf{U}))\] that predicts for all future time slots in parallel.

\subsubsection{Consecutive Dynamic and Semantic Self-Attention}
However, while the simple decoding strategy described above is sufficient to predict future events, it exhibits suboptimal performance by ignoring correlations among concurrent~(semantic) as well as consecutive~(dynamic) events. To address these drawbacks, we apply an additional attentive process on the raw prediction $\hat{\mathbf{Y}}$ to further strengthen the prediction. 

Here, we reuse the Multi-axis Attention Block~(MAB) described previously in a consecutive process.  Concretely, we do the following steps:
\begin{align}
    \begin{split}
        \mathbf{Y}_D &= f_d^{-1}(\textit{MAB}(f_d({\hat{\mathbf{Y}}}))\\
        \mathbf{Y}' = \mathbf{Y}_S &= f_s^{-1}(\textit{MAB}(f_s({\mathbf{Y}_D})),
    \end{split}
\end{align}
where $f_d$ denotes the depth-axis concatenation and $f_s$ denotes the breadth-axis concatenation introduced in Section \ref{sec:encoder}. $f_d ^{-1}$ and $f_s^{-1}$ represent the reverse of the concatenations correspondingly.
\subsection{Model Training}
\label{sec:training}
All of our previous discussions only include an architecture description and do not involve training criteria. In order to successfully capture the semantic and dynamic information of inputs and generalize well, the proposed framework requires consideration of multiple optimization objectives. 

We identify two crucial objectives: 
1) Facilitate capturing time-invariant features (the correlation among crime types) of the semantic encoding $S'$ via contrastive estimation;
2) Make calibrated predictions by maximizing observation likelihood.

\subsubsection{Noise Contrastive Estimation for Semantics Learning}
Contrastive learning~\cite{contrastive} estimates meaningful latent representations by optimizing similarity and difference between pairs of samples. Specifically, contrastive learning encourages the latent representations of observations with similar semantic meaning closer to each other than that of observations from different populations. 
This method is particularly valuable for capturing semantic meaning since the concept of semantic meaning is usually invariant in a dataset. 
The Noise Contrastive Estimation~(NCE)~\cite{NCE} is an invariant of contrastive learning that leverages on data sampling.
In this work, we presume the observation sequence $\mathbf{X}$ maintains its semantic meaning even after shuffling the sequence order. In contrast, different sequences should carry distinct semantic meaning.

While previous works adopt InfoNCE~\cite{CPC,moco}, we found that the dot product based Noise Contrastive Estimation conveys multiple limitations:
\begin{itemize}
    \item Dot product approximates the cosine similarity between two pieces of latent representations without proper regularization; It overlooks the impacts of the representation magnitude. Consequently, the representations identified to be similar could attribute to the magnitude of representation vectors~(or matrix in our case).
    \item During training, the dot product based NCE
    monotonically reduces the value magnitude of the representations. While the reduction does not impact its effectiveness in the self-supervised representation learning tasks~\cite{simclr}, it is harmful in the sequential prediction tasks, where we aim to reconstruct the sequence of observations without introducing magnitude bias that may overfit the training data. We provide concrete evidence for this point in the appendix.
    \item We note the training of the dot product based NCE is slow due to the vanishing gradient problem caused by the product of small numbers. And, InfoNCE tends to need an extended number of training epochs to converge.
\end{itemize}

To mitigate the limitation described above, we propose a Frobenius norm based NCE objective. The proposed objective aims to regularize the latent semantic information $S'$ described in section~\ref{sec:encoder}. Concretely, we define the Frobenius norm based NCE objective as a triplet loss $\mathcal{L}_c$ such that
{\small
\begin{align}
\label{eq:contrastive-derive}
    \begin{split}
        &\mathcal{L}_{c}(S', S'^+, \mathbf{S}'^-)\\
        &= \max\left[- \log \frac{\exp(1/||S'^+-S'||_F^2)}{\sum_{S_i \in \mathbf{S}'^-}\exp(1/||S_i-S'||_F^2)}, 0\right]\\
        &\leq  \max \left[||S'^+-S'||_F^2 - \inf_{S_i \in \mathbf{S}'^-}(||S_i-S'||_F^2)+ \Omega, 0\right],
    \end{split}
\end{align}}
where $\Omega$ is a positive constant, $S'^+$ denotes a representation that hold same semantic meaning of $S'$, and $\mathbf{S}'^-$ denotes a set of representations that have different meaning with $S'$. 

We obtain $S'^+$ by encoding the reordered sequence $\mathbf{X}$ (that produced $S'$ in Equation \ref{eq:MAB_encoding}) and $\mathbf{S}'^-$ by encoding other sequences that $\mathbf{X}^- \neq \mathbf{X}$.

\subsubsection{Maximum Likelihood Estimation for Prediction Accuracy}
To minimize the prediction error~(Empirical Risk Minimization), we adopt an objective function that optimizes both squared Frobenius norm and $\mathcal{L}_1$ norm simultaneously as follows:
\begin{align}
   \begin{split}
        \mathcal{L}_{r} &=  \frac{1}{n}\sum_{\tau=i}^{n}\left [ ||Y_\tau-Y'_\tau||_F^2 +\lambda ||Y_\tau-Y'_\tau||_1\right ]
    \end{split}
    \label{eq:recon-loss}
\end{align}
where $Y_\tau$ and $Y'_\tau$ denotes the ground truth and prediction at time $\tau$.

Finally, we summarize the our loss function as a linear combination of reconstruction loss and contrastive loss for each sequence of training data $(\mathbf{X}, \mathbf{Y})$:
\begin{equation}
    \mathcal{L} = \mathcal{L}_{r}+ \lambda_{c}\mathcal{L}_{c}.
    \label{eq:netloss}
\end{equation}

%% file: content/Experiment.tex
In this section, we perform experiments to answer the following research questions:

\begin{itemize}
    \item RQ1: How does our model compare to traditional and modern state-of-the-art methods of Spatiotemporal event prediction?
    \item RQ2: How effective is contrastive learning and our proposed L2-based contrastive loss?
    \item RQ3: What does the model learn about semantics and dynamics information of crime patterns?
    \item RQ4: How much computation time and parameters does our model require compared to other models?
\end{itemize}

\subsection{Experiment settings}

\subsubsection{Datasets}
In this work, we use the San Francisco (SF) Crime dataset\footnote{https://data.sfgov.org/Public-Safety/Police-Department-Incident-Reports-Historical-2003/tmnf-yvry} and the Vancouver (VAN) Crime dataset\footnote{https://geodash.vpd.ca/opendata/} to evaluate our model. Specifically, we predict $n=3$ weeks of future occurrence maps with $m=5$ weeks of observations. Details of data pre-processing can be found in the appendix.


Note that we aim to predict the number of occurrences instead of if there is an occurrence(i.e. a binary classification task) in the interested space region and time. Therefore models such as ~\cite{deepcrime,Mist,Cross-Interaction} are out of scope of the comparison.

\subsection{Methods}
For baselines, we use:
\begin{itemize}
    \item \textbf{History}: this is a simple baseline for evaluation, which uses the last observed map from the input as prediction.
    \item \textbf{LR} (Linear Regression): assumes a linear relationship between the prediction and historical data of all grids.
    \item \textbf{LSTM}~\cite{lstm}: a baseline time series method with LSTM. 
    \item \textbf{CSAN}~\cite{csan}: a deep neural network model for Spatiotemporal event predictions. It uses VAE for feature encoding and GRU with attention for sequential prediction.
    \item \textbf{ConvGRU}~\cite{convGRU}: a deep neural network that combines convolution operation and GRU for spatio-temporal prediction.
    \item \textbf{NN-CCRF}~\cite{nn-ccrf}: a deep neural network based conditional random field method for crime prediction. It uses a LSTM for temporal pattern capturing and uses a 1-d representation for spatial regions.
    \item \textbf{DuroNet}~\cite{duronet}: an attention-based deep neural network for robust spatiotemporal event prediction. It uses a spatio-temporal convolutional network to calculate the attention parameters and uses a 1-d representation for spatial regions.
\end{itemize}

\subsubsection{Evaluation metrics}
We use two metrics to evaluate the prediction performance.
The averaged Root Mean Square Error (RMSE):
\[RMSE = \sqrt{\frac{1}{n\times h \times w}\sum_{\tau=1}^{n}\sum_{i=1}^{h}\sum_{j=1}^{w}(y_{\tau,i,j}-{y_{\tau,i,j}}')^2} \]
The averaged Mean Absolute Error (MAE):
\[MAE = \frac{1}{n\times h \times w}\sum_{\tau=1}^{n}\sum_{i=1}^{h}\sum_{j=1}^{w}|y_{\tau,i,j}-{y_{\tau,i,j}}'|\]
where $y_{\tau,i,j}$ and ${y_{\tau,i,j}}'$ denotes the ground truth and prediction at time $t$ and location $(i,j)$ .

\subsection{RQ1: Performance Comparison}
\label{sec:performance comparison}

\begin{table}[htbp]
\caption{Performance comparison on San Francisco crime data. Lower is better.}
\centering
\resizebox{\linewidth}{!}{
\begin{tabular}{ccccccccc}
\toprule
\multirow{2}{*}{Model} & \multicolumn{2}{l}{Theft} & \multicolumn{2}{l}{Other} & \multicolumn{2}{l}{Non-criminal(NC)} & \multicolumn{2}{l}{Assault} \\ \cmidrule(l){2-9} 
 &
  \multicolumn{1}{l}{RMSE} &
  \multicolumn{1}{l}{MAE} &
  \multicolumn{1}{l}{RMSE} &
  \multicolumn{1}{l}{MAE} &
  \multicolumn{1}{l}{RMSE} &
  \multicolumn{1}{l}{MAE} &
  \multicolumn{1}{l}{RMSE} &
  \multicolumn{1}{l}{MAE} \\ \midrule
\multicolumn{1}{l}{History} &
  \multicolumn{1}{l}{3.8659} &
  \multicolumn{1}{l}{2.3255} &
  \multicolumn{1}{l}{3.6223} &
  \multicolumn{1}{l}{2.1028} &
  \multicolumn{1}{l}{2.7176} &
  \multicolumn{1}{l}{1.6186} &
  \multicolumn{1}{l}{2.5853} &
  \multicolumn{1}{l}{1.5378} \\ \midrule
  
  \multicolumn{1}{l}{LR} &
  \multicolumn{1}{l}{3.7907} &
  \multicolumn{1}{l}{2.3909} &
  \multicolumn{1}{l}{3.5549} &
  \multicolumn{1}{l}{2.1180} &
  \multicolumn{1}{l}{2.5825} &
  \multicolumn{1}{l}{1.6371} &
  \multicolumn{1}{l}{2.4105} &
  \multicolumn{1}{l}{1.5768} \\ \midrule
  
  \multicolumn{1}{l}{CSAN} &
  \multicolumn{1}{l}{4.2242} &
  \multicolumn{1}{l}{2.3763} &
  \multicolumn{1}{l}{4.1250} &
  \multicolumn{1}{l}{1.9898} &
  \multicolumn{1}{l}{2.4885} &
  \multicolumn{1}{l}{1.4219} &
  \multicolumn{1}{l}{1.9753} &
  \multicolumn{1}{l}{\textbf{1.1953}} \\ \midrule
  
  \multicolumn{1}{l}{LSTM} &
  \multicolumn{1}{l}{3.6369} &
  \multicolumn{1}{l}{2.0869} &
  \multicolumn{1}{l}{3.5418} &
  \multicolumn{1}{l}{1.8289} &
  \multicolumn{1}{l}{2.4098} &
  \multicolumn{1}{l}{1.3855} &
  \multicolumn{1}{l}{2.0929} &
  \multicolumn{1}{l}{1.2520} \\ \midrule
  
  \multicolumn{1}{l}{ConvGRU} &
  \multicolumn{1}{l}{3.9442} &
  \multicolumn{1}{l}{2.3269} &
  \multicolumn{1}{l}{3.6555} &
  \multicolumn{1}{l}{2.0229} &
  \multicolumn{1}{l}{2.3719} &
  \multicolumn{1}{l}{1.4229} &
  \multicolumn{1}{l}{2.1552} &
  \multicolumn{1}{l}{1.3054} \\ \midrule
  
  \multicolumn{1}{l}{NN-CCRF} &
  \multicolumn{1}{l}{3.5659} &
  \multicolumn{1}{l}{2.1056} &
  \multicolumn{1}{l}{3.5899} &
  \multicolumn{1}{l}{1.9050} &
  \multicolumn{1}{l}{2.3777} &
  \multicolumn{1}{l}{1.3697} &
  \multicolumn{1}{l}{1.9640} &
  \multicolumn{1}{l}{1.2349} \\ \midrule

  \multicolumn{1}{l}{DuroNet} &
  \multicolumn{1}{l}{3.4350} &
  \multicolumn{1}{l}{2.0036} &
  \multicolumn{1}{l}{3.3111} &
  \multicolumn{1}{l}{1.7562} &
  \multicolumn{1}{l}{2.2884} &
  \multicolumn{1}{l}{1.3340} &
  \multicolumn{1}{l}{\textbf{1.9277}} &
  \multicolumn{1}{l}{1.2055} \\ \midrule

  \multicolumn{1}{l}{MAPSED} &
  \multicolumn{1}{l}{\textbf{3.3738}} &
  \multicolumn{1}{l}{\textbf{2.0281}} &
  \multicolumn{1}{l}{\textbf{3.1550}} &
  \multicolumn{1}{l}{\textbf{1.7444}} &
  \multicolumn{1}{l}{\textbf{2.2107}} &
  \multicolumn{1}{l}{\textbf{1.3268}} &
  \multicolumn{1}{l}{1.9892} &
  \multicolumn{1}{l}{1.2055} \\ \bottomrule
\end{tabular}}
\label{tab:SF}
\end{table}

\begin{table}[htbp]
\caption{Performance comparison on Vancouver crime data. Lower is better.}
\centering
\resizebox{\linewidth}{!}{
\begin{tabular}{ccccccccc}

\toprule
\multirow{2}{*}{Model} & \multicolumn{2}{l}{Vehicle theft} & \multicolumn{2}{l}{Mischief} & \multicolumn{2}{l}{Burglary} & \multicolumn{2}{l}{Personal offence} \\ \cmidrule(l){2-9} 
 &
  \multicolumn{1}{l}{RMSE} &
  \multicolumn{1}{l}{MAE} &
  \multicolumn{1}{l}{RMSE} &
  \multicolumn{1}{l}{MAE} &
  \multicolumn{1}{l}{RMSE} &
  \multicolumn{1}{l}{MAE} &
  \multicolumn{1}{l}{RMSE} &
  \multicolumn{1}{l}{MAE} \\ \midrule
  
 \multicolumn{1}{l}{History} &
  \multicolumn{1}{l}{2.8065} &
  \multicolumn{1}{l}{1.5590} &
  \multicolumn{1}{l}{1.6305} &
  \multicolumn{1}{l}{0.8602} &
  \multicolumn{1}{l}{1.2587} &
  \multicolumn{1}{l}{0.7751} &
  \multicolumn{1}{l}{1.4423} &
  \multicolumn{1}{l}{0.5690} \\ \midrule
  
    \multicolumn{1}{l}{LR} &
  \multicolumn{1}{l}{2.8199} &
  \multicolumn{1}{l}{1.6503} &
  \multicolumn{1}{l}{1.7979} &
  \multicolumn{1}{l}{0.9760} &
  \multicolumn{1}{l}{1.2874} &
  \multicolumn{1}{l}{0.9184} &
  \multicolumn{1}{l}{1.3688} &
  \multicolumn{1}{l}{0.6480} \\ \midrule
  
  \multicolumn{1}{l}{CSAN} &
  \multicolumn{1}{l}{2.9763} &
  \multicolumn{1}{l}{1.4701} &
  \multicolumn{1}{l}{1.2522} &
  \multicolumn{1}{l}{0.7185} &
  \multicolumn{1}{l}{1.0295} &
  \multicolumn{1}{l}{0.7139} &
  \multicolumn{1}{l}{1.4181} &
  \multicolumn{1}{l}{0.5608} \\ \midrule
  
  \multicolumn{1}{l}{LSTM} &
  \multicolumn{1}{l}{2.8291} &
  \multicolumn{1}{l}{1.4796} &
  \multicolumn{1}{l}{1.4606} &
  \multicolumn{1}{l}{0.8017} &
  \multicolumn{1}{l}{1.1294} &
  \multicolumn{1}{l}{0.7726} &
  \multicolumn{1}{l}{1.4593} &
  \multicolumn{1}{l}{0.5950} \\ \midrule
  
  \multicolumn{1}{l}{ConvGRU} &
  \multicolumn{1}{l}{3.0047} &
  \multicolumn{1}{l}{1.5878} &
  \multicolumn{1}{l}{1.3633} &
  \multicolumn{1}{l}{0.7743} &
  \multicolumn{1}{l}{1.0569} &
  \multicolumn{1}{l}{0.7293} &
  \multicolumn{1}{l}{1.4832} &
\multicolumn{1}{l}{0.5666} \\ \midrule
  
  \multicolumn{1}{l}{NN-CCRF} &
  \multicolumn{1}{l}{2.6353} &
  \multicolumn{1}{l}{1.5428} &
  \multicolumn{1}{l}{1.2814} &
  \multicolumn{1}{l}{0.7713} &
  \multicolumn{1}{l}{1.0089} &
  \multicolumn{1}{l}{0.7116} &
  \multicolumn{1}{l}{1.3931} &
  \multicolumn{1}{l}{0.6323} \\ \midrule
  
  \multicolumn{1}{l}{DuroNet} &
  \multicolumn{1}{l}{\textbf{2.4970}} &
  \multicolumn{1}{l}{1.3677} &
  \multicolumn{1}{l}{1.2627} &
  \multicolumn{1}{l}{0.7421} &
  \multicolumn{1}{l}{0.9882} &
  \multicolumn{1}{l}{0.6992} &
  \multicolumn{1}{l}{1.3327} &
  \multicolumn{1}{l}{0.5578} \\ \midrule
  
  \multicolumn{1}{l}{MAPSED} &
  \multicolumn{1}{l}{2.5190} &
  \multicolumn{1}{l}{\textbf{1.3566}} &
  \multicolumn{1}{l}{\textbf{1.2575}} &
  \multicolumn{1}{l}{\textbf{0.7233}} &
  \multicolumn{1}{l}{\textbf{0.9845}} &
  \multicolumn{1}{l}{\textbf{0.6763}} &
  \multicolumn{1}{l}{\textbf{1.2418}} &
  \multicolumn{1}{l}{\textbf{0.5208}} \\ \bottomrule
  
\end{tabular}}
\label{tab:Vancouver}
\end{table}

The performance comparisons over San Francisco crime data and Vancouver Crime data are displayed in Table \ref{tab:SF} and Table \ref{tab:Vancouver}. We observe that, among all approaches, the proposed MAPSED model shows significantly better testing performance in both tasks. In particular, on the San Francisco crime data, we note some of the existing models do not even outperform the baseline approach, which directly copies the last observation as its prediction. This observation reflects our intuition about the overfitting of the existing methods.

\begin{figure}[t]
    \centering
    \subfigure[Prediction]{\includegraphics[width=0.598\linewidth]{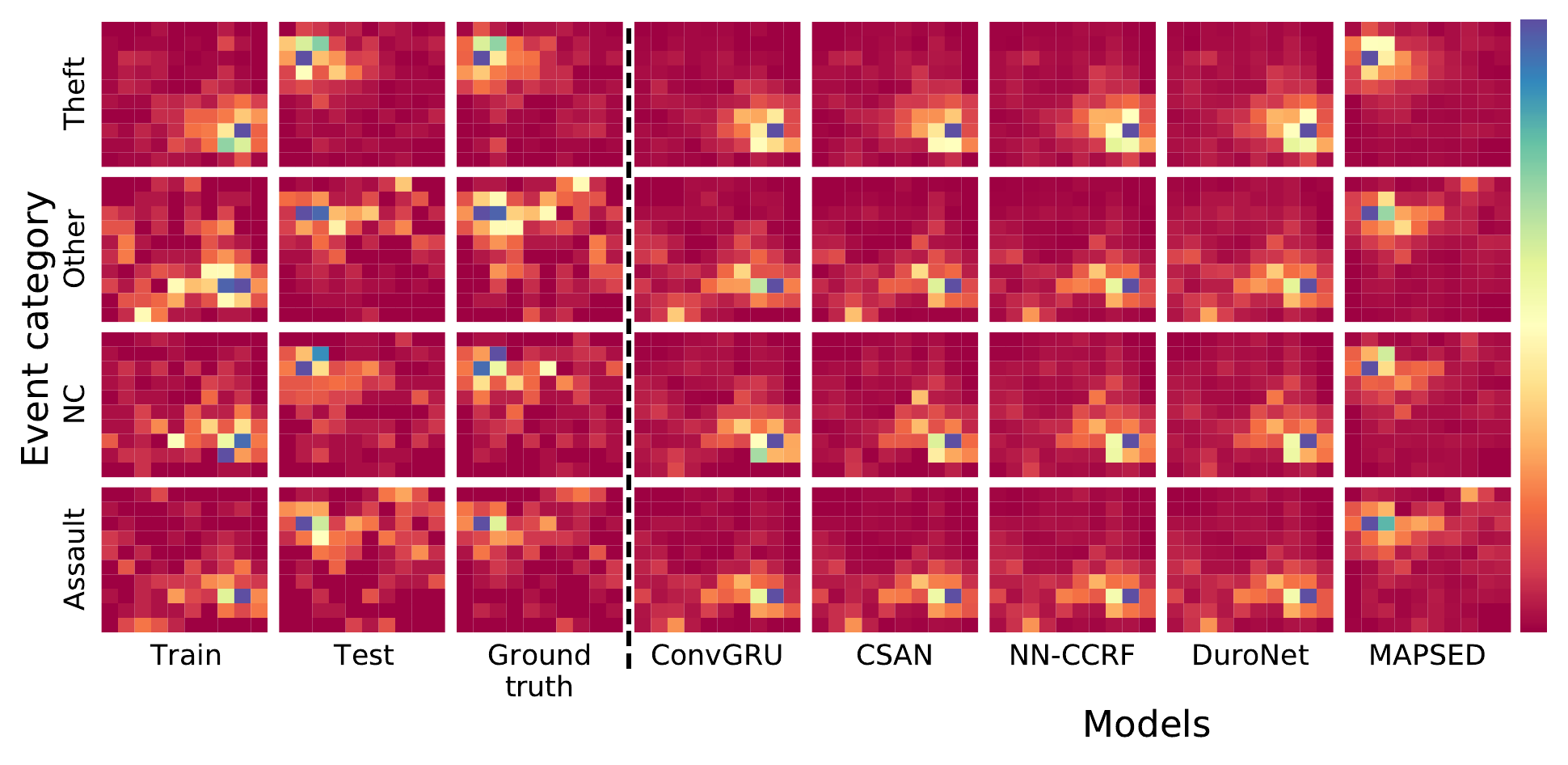}}
    \subfigure[MAE]{\includegraphics[width=0.388\linewidth]{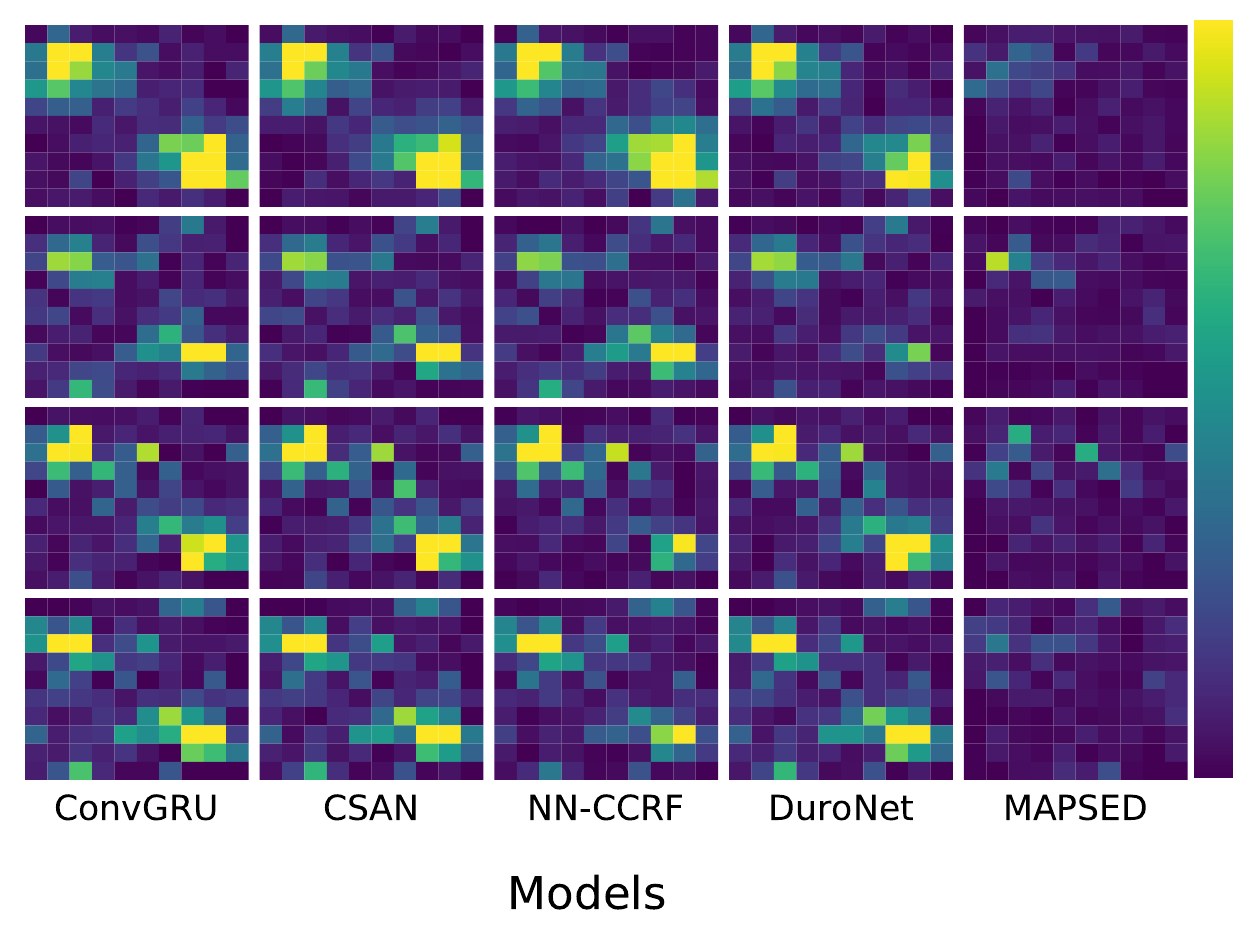}}
    \caption{Prediction and MAE visualization on SF crime dataset with rotated testing data. Lower is better. The testing data used is a rotation of 180 degrees from the original observation. A typical training input is shown on the first column from the left in (a).}
    \label{fig:rotated-input}
\end{figure}

\begin{figure}[t]
    \centering
    \subfigure[Prediction]{\includegraphics[width=0.598\linewidth]{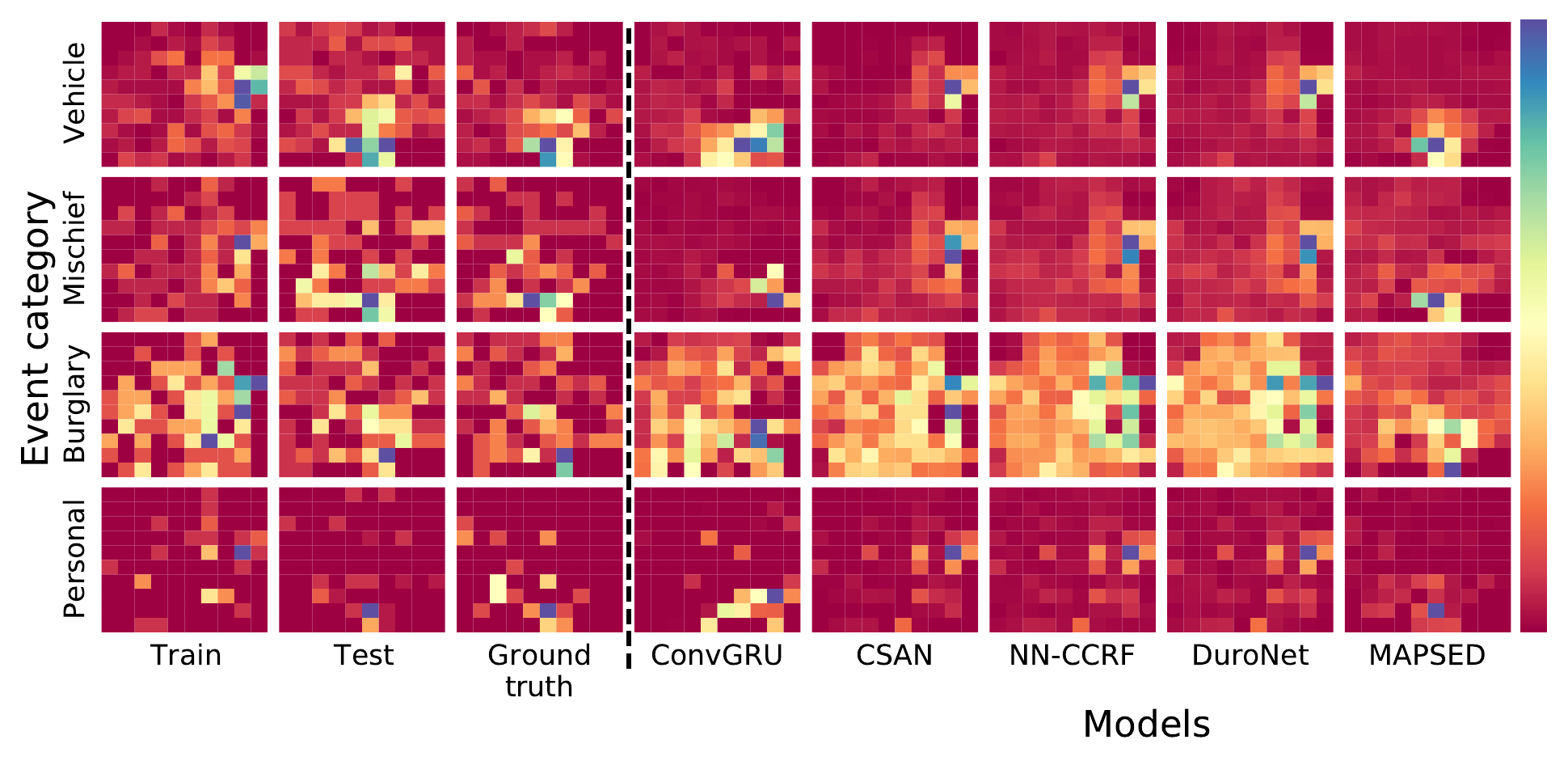}}
    \subfigure[MAE]{\includegraphics[width=0.388\linewidth]{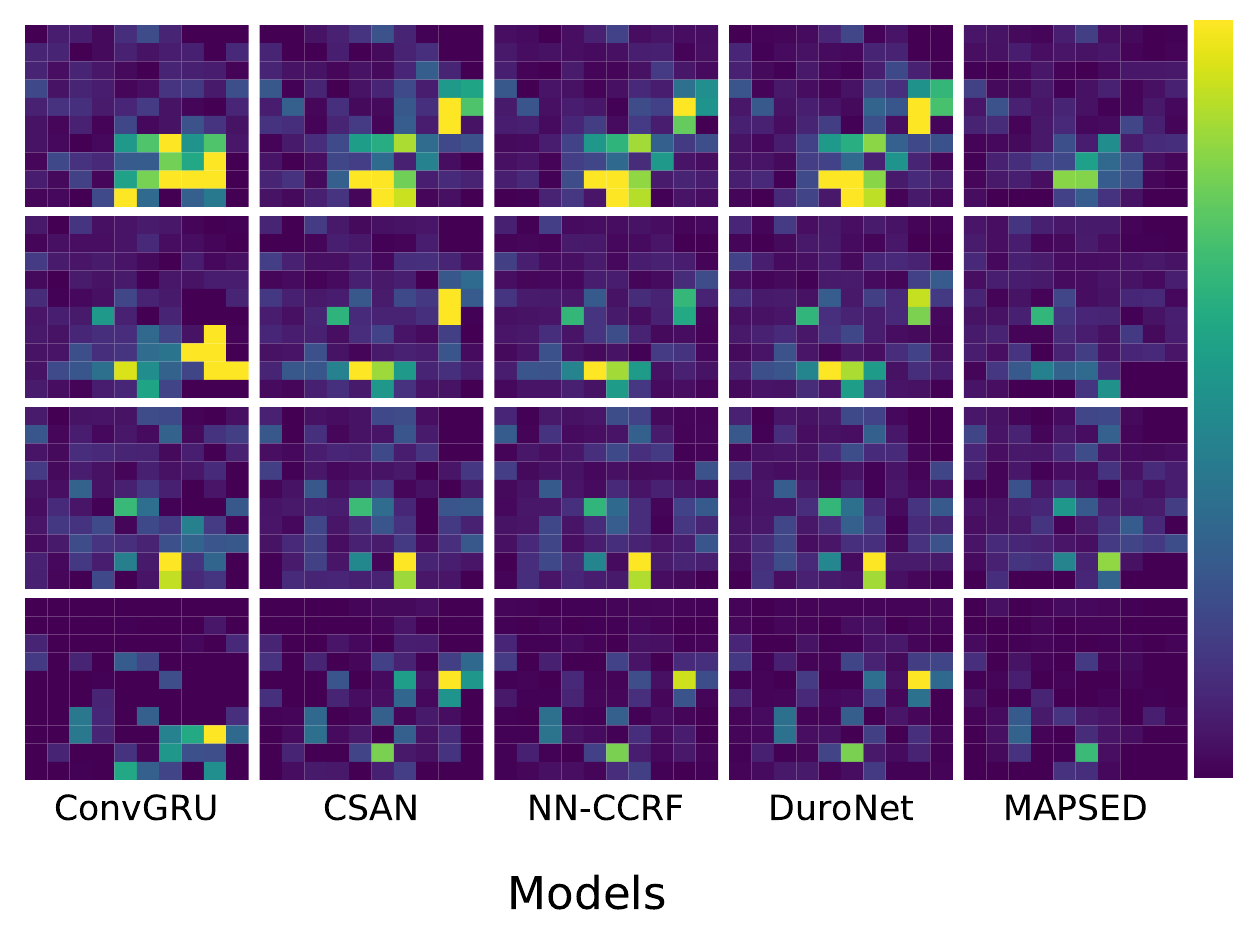}}
    \caption{Prediction and MAE visualization on VAN crime dataset with rotated testing data. The testing data used is a rotation of 90 degrees to the original observation. A typical training input is shown on the first column from the left in (a).}
    \label{fig:VAN-rotated-input}
\end{figure}

To further investigate MAPSED and the spatio-temporal event prediction baseline models' generalization ability, we tested these models with a test sequence rotation. Figure~\ref{fig:rotated-input} and \ref{fig:VAN-rotated-input} show the resulting output and MAE of this experiment. Interestingly, we observed that all spatio-temporal event prediction baseline models(ConvGRU, CSAN, NN-CCRF, and DuroNet) generate predictions similar to the typical training data with the bottom right area as the crime hotspots, which shows that the prediction completely ignores the observed test sequences and strongly overfits the typical training data observations. In contrast, the proposed MAPSED method could correctly identify the location of events without overfitting. This observation justifies our previous hypothesis that MAPSED is more suitable for the tasks with sparse observations in combating overfitting. 

\subsection{RQ2: Effect of contrastive learning}
\begin{figure}[tbp]
\centering
\includegraphics[width=0.95\linewidth]{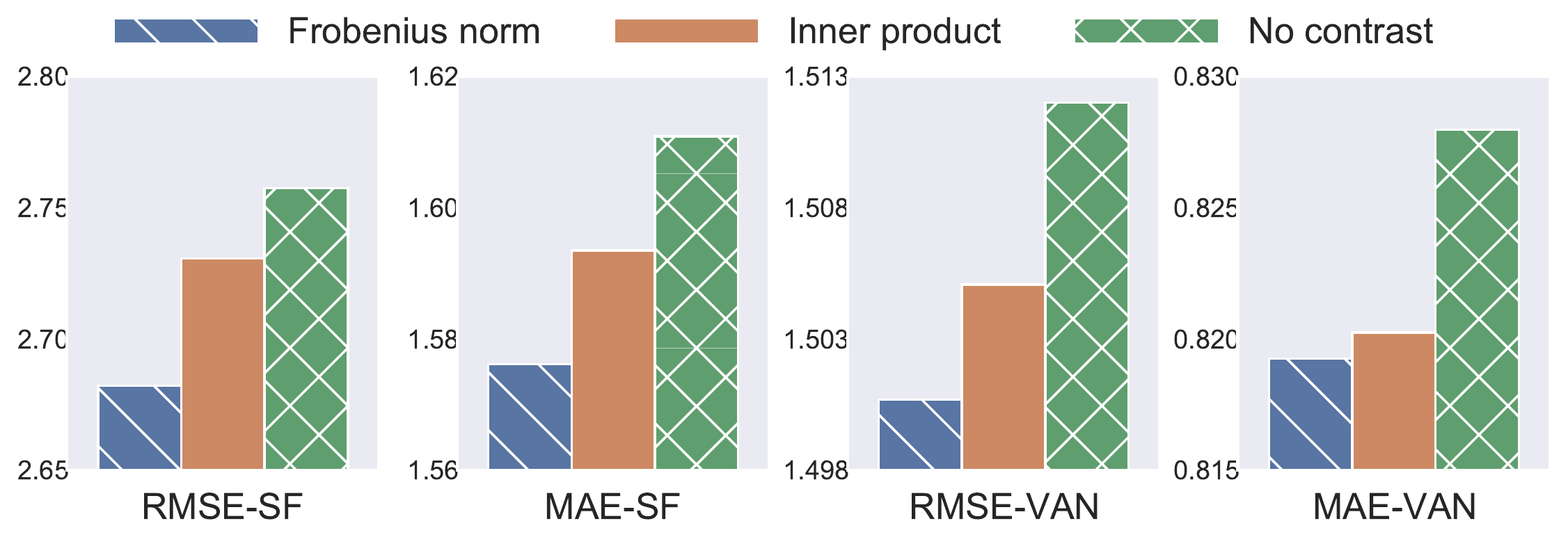}
\caption{Performance comparison between contrastive implementations. The mean errors across four categories are reported in each plot.}
\label{fig:contrast}
\end{figure}

To examine the effectiveness of our proposed Frobenius norm based contrastive loss, we evaluated the performance of MAPSED with an inner-product contrastive loss and no contrastive loss. As shown in Figure \ref{fig:contrast}, the Frobenius norm based contrastive model has the lowest errors across both evaluations and datasets. Furthermore, incorporating contrastive learning with both implementations helps boost the performance of MAPSED. 

\begin{figure}[htbp]
    \centering
    \includegraphics[width=0.95\linewidth]{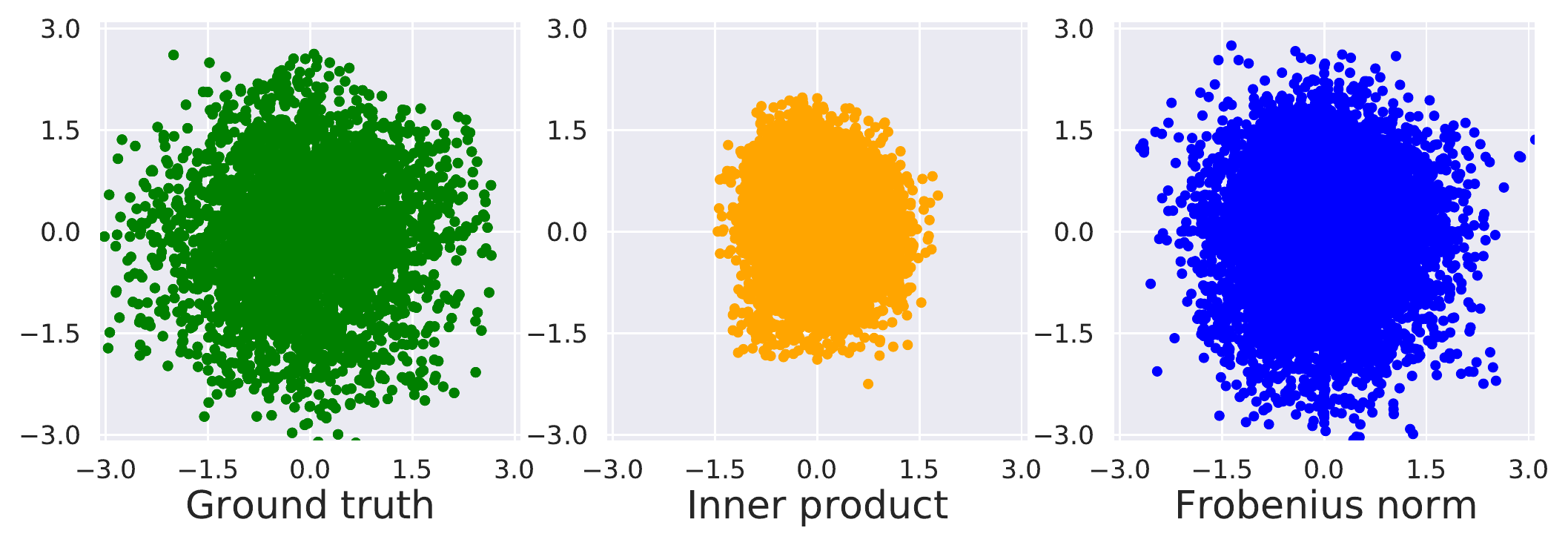}
    \caption{Latent value visualization of ground truth, inner product InfoNCE, and our proposed Frobenius norm based InfoNCE.}
    \label{fig:contrast-softmax}
\end{figure}

Figure \ref{fig:contrast-softmax} provides additional information about the difference between inner product-based and Frobenius norm-based contrastive objectives in the 2D latent representation space (as part of representation learning). We note the inner product based contrastive learning would monotonously reduce the latent representation magnitude. While such property does not harm the downstream classification tasks' performance, it may introduce additional difficulty to the reconstruction tasks as we need in Spatiotemporal event prediction.

\subsection{RQ3: Dynamics and Semantics exploration}
\label{sec:semantics and dynamics experiment}
To investigate the semantics and dynamics information learned from the observations, we conduct the two experiments described below.
\subsubsection{Semantics}
\begin{figure}[tbp!]
\centering
\subfigure[SF]{\includegraphics[width=0.49 \linewidth]{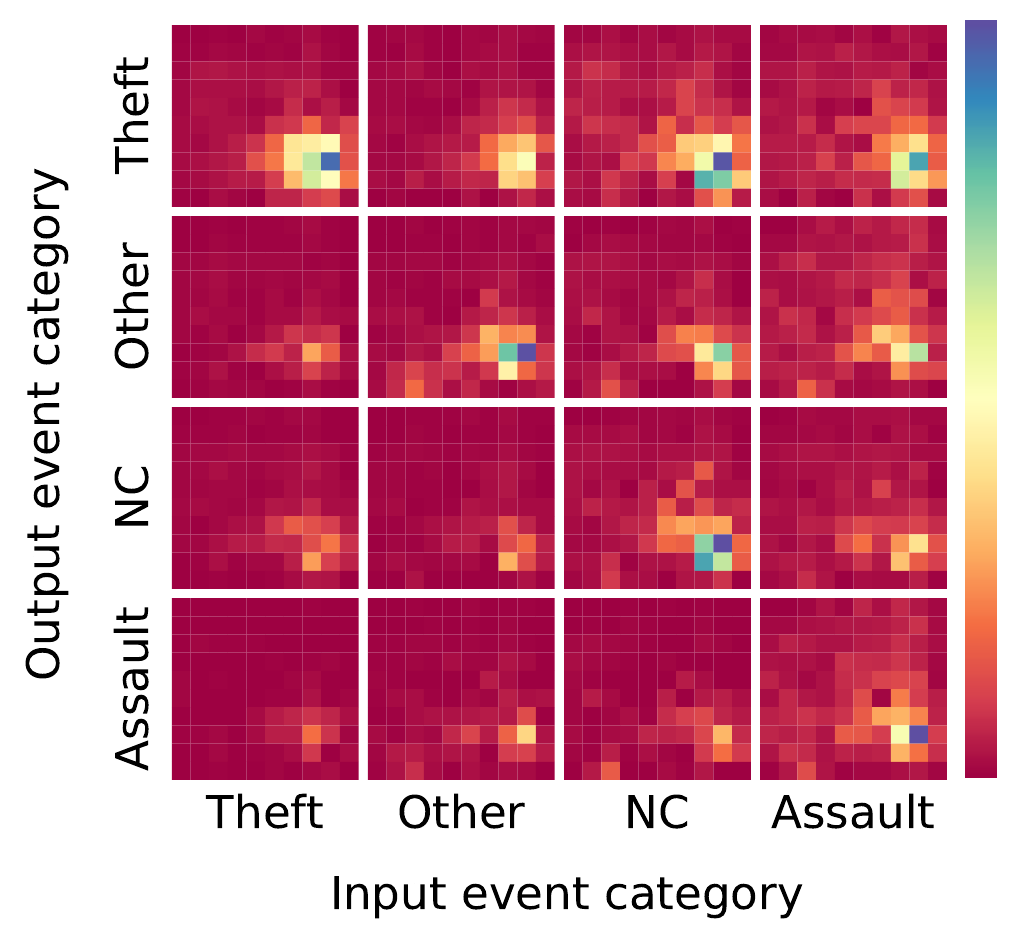}}
\subfigure[VAN]{\includegraphics[width=0.49 \linewidth]{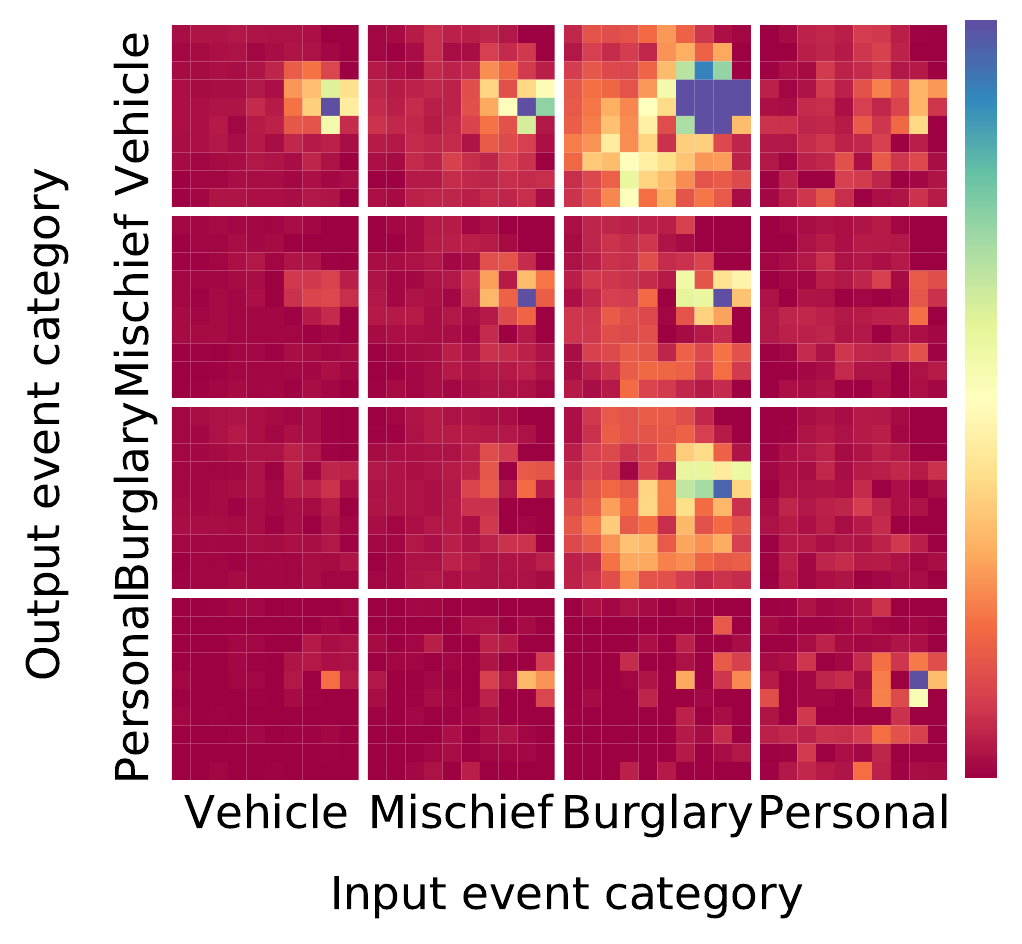}}
\caption{The prediction on SF and VAN crime data with all crime events aggregated into one of the four categories. We use this experiment to explore the semantics learned by MAPSED.}
\label{fig:semantics}
\end{figure}

To explore the semantics information learned by MAPSED, we aggregate all crime events into one category in the historical observation. The prediction from the trained MAPSED model is shown in Figure \ref{fig:semantics}. 

It can be observed that the prediction includes events in all four categories, indicating the categorical correlations learned by MAPSED. For example, in  Figure \ref{fig:semantics} (a), when crimes of category non-criminal (NC) happens, theft also happens in similar areas. This can be explained by the fact that  $32.7\%$ and of the NC events are associated with the description "Lost property". Similarly, we observe a strong correlation between Burglary and Vehicle theft. This corresponds to the perspective from crime research that motor vehicles are often stolen for transporting people to other crime scenes, especially burglaries\cite{crimereport}. According to statistics Canada \cite{crime2007}, another common purpose for stolen vehicles is joy-riding. This type of vehicle theft crimes usually highly correlates with mischief; incidentally, the offenders in these cases are mostly teenagers indicating the semantic connections between these two crime categories.

In summary, this experiment demonstrates that MAPSED effectively learns how the spatiotemporal occurrences of a particular event type affect other events. In other words, MAPSED appears to effectively capture meaningful semantic information underlying the crime predictions.

\subsubsection{Dynamics}
\begin{figure}[htbp]
    \centering
    \includegraphics[width=0.8\linewidth]{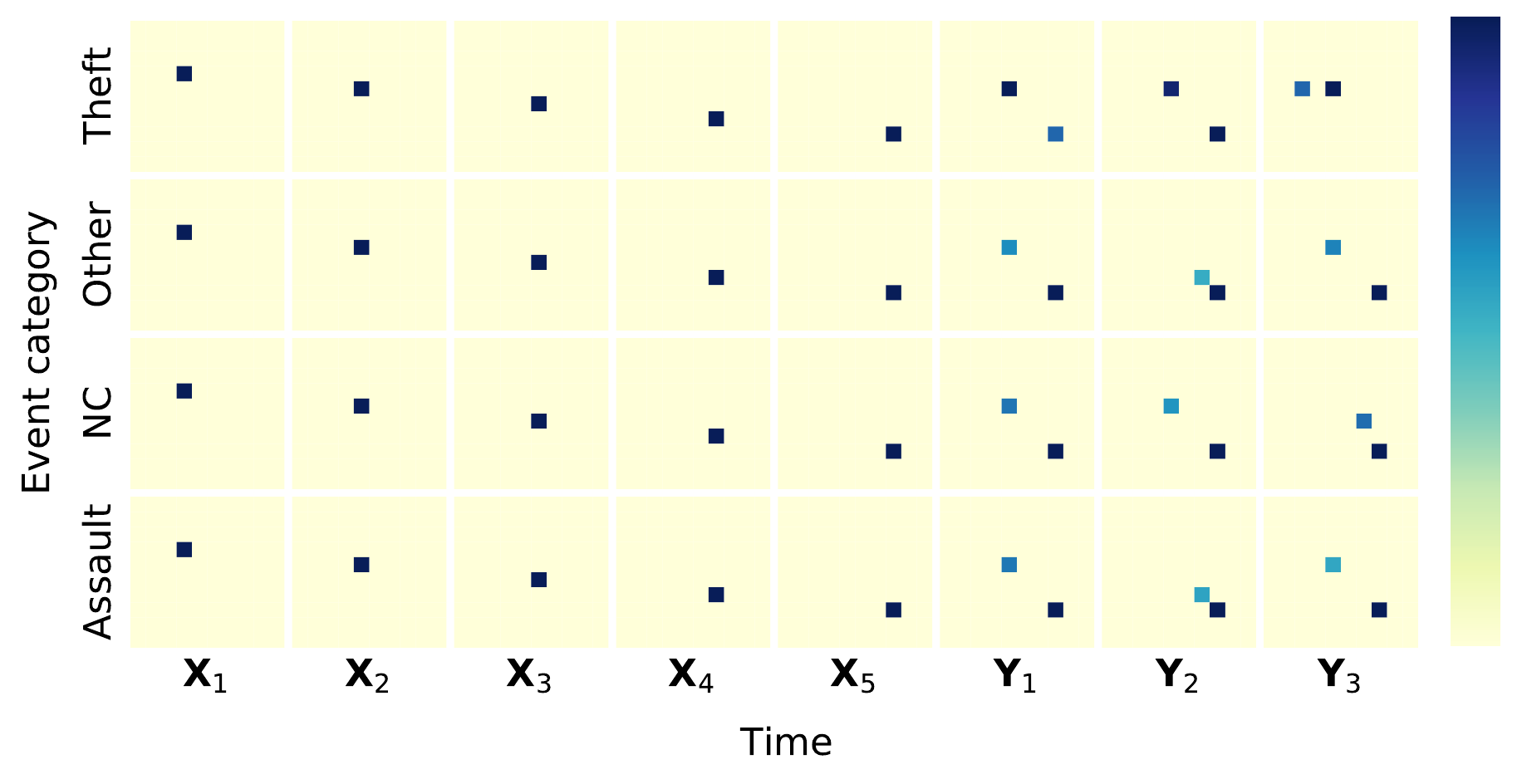}
    \caption{The input and output of dynamics exploration experiment with SF crime data. The input crimes are aggregated into one spatial grid and move along the diagonal line to the bottom-right corner. The First five columns are observations, and the last three columns are predictions.}
    \label{fig:dynamics}
\end{figure}
To explore the dynamics information learned by the trained MAPSED, we aggregate all events of each type into a single location that moves diagonally towards the bottom right corner with time. Figure \ref{fig:dynamics} shows the input and output of the experiment. In general, the predictions also tend to have crime hotspots along the diagonal line with the trend of moving to the bottom-right corner.

Overall, this experiment demonstrates that MAPSED captures how crime events in a spatial region may propagate through time. Therefore, we can claim that MAPSED learns useful dynamics for the crime predictions.
\subsection{RQ4: Resource consumption comparison}

\begin{figure}[tbp]
\centering
\subfigure[Training phase]{\includegraphics[width=0.49\linewidth]{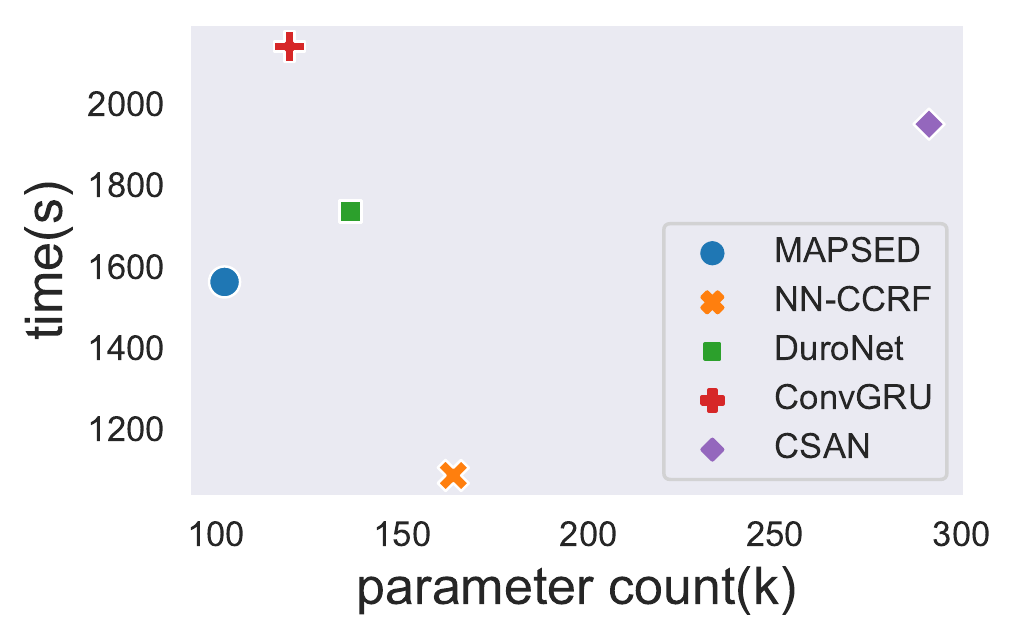}}
\subfigure[Inferencing phase]{\includegraphics[width=0.49\linewidth]{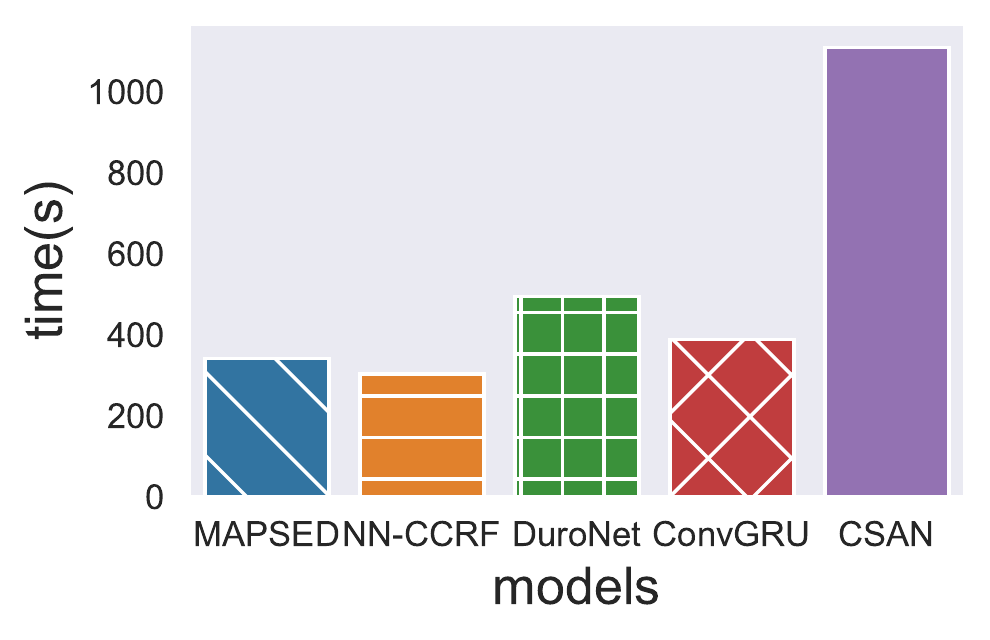}}
\caption{The analysis of resource consumption of CSAN, ConvGRU, and MAPSED. Lower is better.}
\label{fig:resource}
\end{figure}

Figure \ref{fig:resource} shows the resource consumption comparison among the candidate spatiotemporal prediction models. In training phase (a), MAPSED uses fewer parameters than the baseline spatio-temporal event prediction models and has less training time than most baseline models(all except NN-CCRF). In inference phase(b), we observe that MAPSED either requires comparable compuation time(NN-CCRF) or requires less computation time to predict.


%% file: content/Conclusion.tex
In this paper, we proposed MAPSED, a multi-axis attention model for spatiotemporal event prediction with sparse data. Specifically, we adopted a tensor-centric method that simultaneously learned dependencies over space, time and event categories. In addition, we proposed a novel Frobenius norm based contrastive loss to improve latent representation learning.  We validated MAPSED on two publicly accessible urban crime datasets and showed that MAPSED outperforms both classical and state-of-the-art deep learning models, accurately captures dynamics, and uses less parameters and comparable or less computation time than competing state-of-the-art models.

%% file: appendix.tex
\appendix

\subsection{Derivation of the proposed Frobenius norm based contrastive loss}
We derive our proposed Frobenius norm based contrastive loss as follows:
{\small
\begin{align}
\label{eq:contrastive-derive-detail}
    \begin{split}
        &\mathcal{L}_{c}(S', S'^+, \mathbf{S}'^-)\\
        =& \max\left[- \log \frac{\exp(1/||S'^+-S'||_F^2)}{\sum_{S_i \in \mathbf{S}'^-}\exp(1/||S_i-S'||_F^2)}, 0\right]\\
        =& \max\left[||S'^+-S'||_F^2 +\log\sum_{S_i \in \mathbf{S}'^-}\exp(1/||S_i-S'||_F^2), 0 \right]\\
        \leq&  \max \left[||S'^+-S'||_F^2 + \sup_{S_i \in \mathbf{S}'^-}(\frac{1}{||S_i-S'||_F^2})+ \Omega, 0\right]\\
        =&  \max \left[||S'^+-S'||_F^2 - \inf_{S_i \in \mathbf{S}'^-}(||S_i-S'||_F^2)+ \Omega, 0\right]
    \end{split}
\end{align}}

\subsection{Scalability of the Proposed Model}
As described previously, the proposed MAPSED model does not reduce the observation dimension during its training and inference time, which may limit its application in practice. 

This section describes how we scale up MAPSED on real applications by reducing and re-scaling observation dimensions in real-time.

Concretely, we propose two options: Pooling/unpooling and Variational Auto-encoders~(VAEs):
\begin{itemize}
    \item {\bf Pooling for dimension reduction:}
Pooling is a popular method to down-sample the input representation into a lower dimensional space by extracting the most informative features from local sub-regions. To reconstruct the down-sampled representation back to its original dimension, we need either remember the feature exaction coordinates~\cite{noh2015learning} or copying the same value to all sub-region.

\item {\bf VAE for dimension reduction:}
Auto-encoders are well-known as an alternative dimension reduction method. However, as the deterministic auto-encoder cannot reconstruct meaningful observation after latent representation perturbation, it is not favorable since we aim to reconstruct the raw observations from the latent predictions. To this end, we propose to use Variational Auto-encoder~\cite{vae}. As the VAE regularizes its latent representation through KL-divergence, it can tolerate small prediction noises. 
\end{itemize}

In our experiment, we pre-train a VAE to reduce and re-scale the observations/predictions' dimension at each timestamp.

\subsection{Training algorithm of the Proposed Model}
To facilitate reproducibility, we summarize the architecture description along with the objectives into an Algorithm~\ref{alg:algorithm}.

\begin{algorithm}[tb]
\DontPrintSemicolon
\SetAlgoLined
\SetKwInOut{Input}{Input}\SetKwInOut{Output}{Output}
\SetKwRepeat{repeat}{repeat}{until}
\Input{Historical observations $\mathbf{X} \in R^{m\times c\times h \times w}$; \newline Ground truth of future $\mathbf{Y} \in R^{n\times c\times h \times w}$; \newline
The pre-trained vae $g$ with the encoder $g_{enc}$ and the decoder $g_{dec}$.}

\tcp*[h]{construct positive samples}\;
$\mathbf{X}^+ \leftarrow$ random permutation of $\mathbf{X}$\;
\tcp*[h]{construct negative samples}\;
Randomly select a set of negative samples $\{\mathbf{X}^-_i\}$ from the dataset , where $\mathbf{X}^-_i \neq \mathbf{X}$.\;
Initialize all parameters in MAPSED \;
\repeat{stopping criterion met}{
$\mathbf{X'} =g_{enc}(\mathbf{X})$ , $\mathbf{{X^+}'} = g_{enc}(\mathbf{X^+})$\;
Initialize $(D$, $S)$ with  $\mathbf{X'} $ \;
Initialize  $S^+$ with $\mathbf{{X^+}'}$ \;

$D',S',\mathbf{U} \leftarrow  \mathrm{Encoder}(D, S, \mathbf{X} )$\;
${S'}^+ \leftarrow $ Semantics Extractions$(S^+)$\;
$\mathbf{S'}^- \leftarrow \{$Semantics Extractions$(\mathbf{X}^-_i)\}$\;
$\mathbf{Y'}  \leftarrow $
$g_{dec}(\mathrm{Decoder}(\mathbf{U}))$\;
Compute $\mathcal{L} = \mathcal{L}_{r}+ \lambda_{c}\mathcal{L}_{c}$ with $S', {S'}^+, \mathbf{S'}^-, \mathbf{Y}, \mathbf{Y'}$\;
Update all parameters \textit{w.r.t} $\mathcal{L}$
}
\caption{The training algorithm of MAPSED}
\label{alg:algorithm}
\end{algorithm}

\subsection{Detailed data preprocessing}
The San Francisco crime dataset ranges from January 1st 2003 to May 15th, 2018 and the Vancouver crime dataset ranges from January 1st 2003 to July 1st, 2017. Each crime record contains information of geospatial location, timestamp of occurrence, and type of crimes. 

To create the spatial maps, we divide each city into $10 \times 10$ grids of the same size and aggregate event occurrences into the grids. As for crime categories, we pick the most common 4 categories of crimes in our experiment same as in \cite{csan}. We aggregate time into weekly intervals and predict the crime occurrences of the future $n=3$ weeks given the historical $m=5$ weeks. To generate more sequences from the collected data, we adopt a sliding window technique. Specifically, to avoid data leakage, we firstly partitioned the entire time range into non-overlapping sets of periods and assign them separately to train, validation and test periods. Then, in the selected periods, we generated the consecutive event sequences $\{\textbf{X},\textbf{Y}\}$ with a sliding window of 8 weeks. Finally, for the San Francisco dataset, the train-test split is about 5:1. For the Vancouver dataset, the train-test split is about 3:1.

\subsection{Data Augmentation}
We apply data augmentation during training to enlarge the dataset. Specifically, in training phase, we augment each training sequence  $\{\textbf{X}, \textbf{Y}\}$ by randomly flipping each $X_{\tau}$ and $Y_{\tau}$ horizontally with a $50\%$ chance followed by a rotation with an angle of randomly chosen from 0, 90, 180, and 270 degrees.

\subsection{Details of the baseline implementations}

Here, we list the detail parameter settings for the baselines:
\begin{itemize}
    \item \textbf{LSTM}~\cite{lstm}: We train one LSTM for each event category with a hidden dimension of 30 and rnn layers of 3.
    \item \textbf{CSAN}~\cite{csan}: the parameter settings are kept the same as in the original paper with $z=6$.
    \item \textbf{ConvGRU}~\cite{convGRU}: we implemented the encoder-decoder 4-layer network with hidden dimension of $(16,10,10)$.
    \item \textbf{NN-CCRF}~\cite{nn-ccrf}: we train one NN-CCRF models for each event category. Each model has a  hidden dimension of 25 with 4-layer rnn.
    \item \textbf{DuroNet}~\cite{duronet}: Similar to NN-CCRF, we train 4 duronets. For each one, we use 2 layers, 2 attention heads, key and value vector of dimension 6, the inner layer feed-forward dimension of 6, the temporal convolution kernel of size 3, and the spatial convolution kernel of size 2.
\end{itemize}

\subsection{Additional Experiment Results}

\begin{figure}[tbp]
    \centering
    \includegraphics[width=0.95\linewidth]{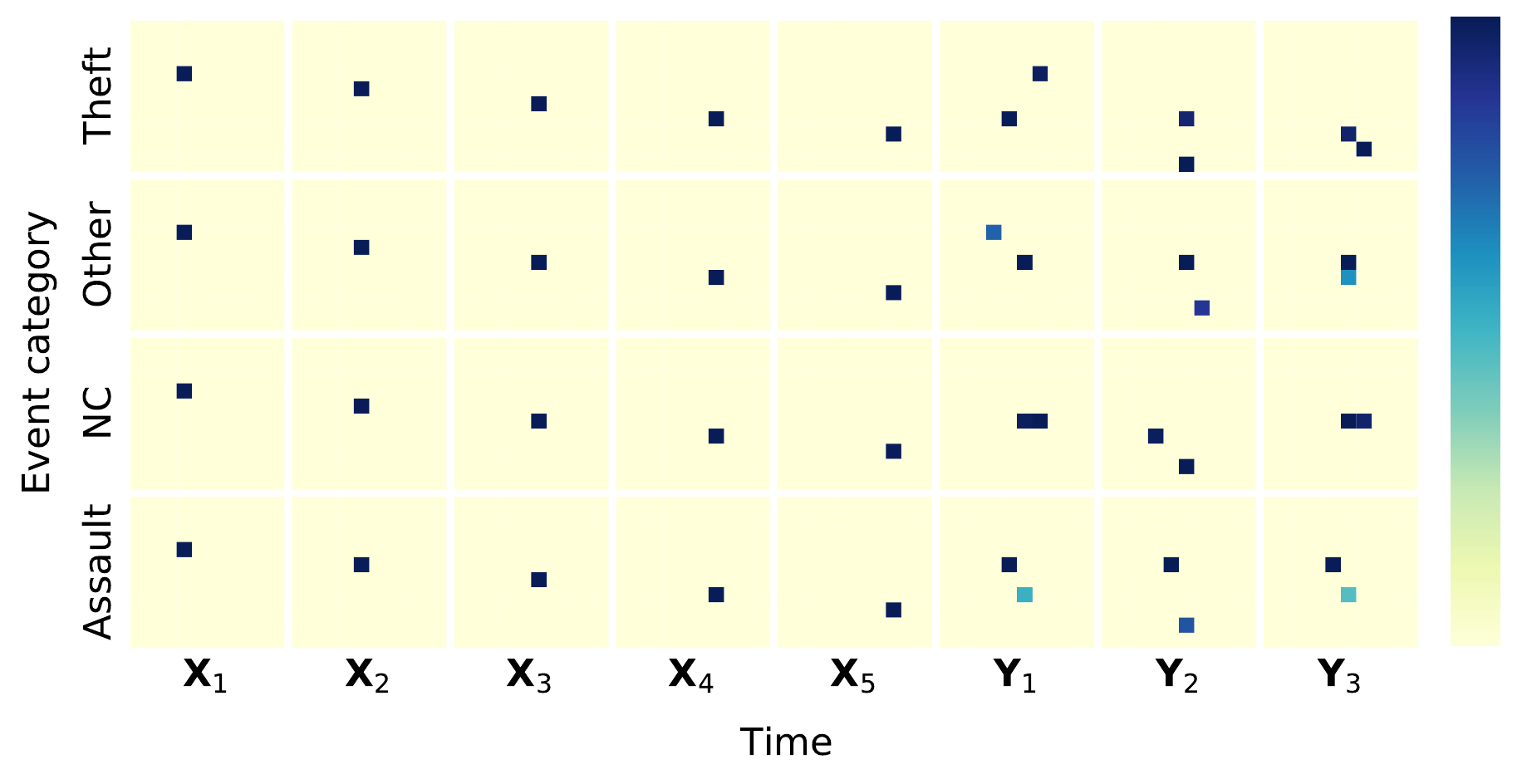}
    \caption{The input and output of dynamics exploration experiment with VAN crime data. The input crimes are aggregated into one spatial grid and moves along the diagonal line to the bottom-right corner. First 5 columns are observations, and the last 3 columns are predictions.}
    \label{fig:VAN-dynamics}
\end{figure}

\noindent \textbf{Dynamics exploration on VAN crime data}\\
In the main body of the paper, we explored the learned dynamics by aggregating all observed crimes onto a single grid moving along the diagonal line overtime with SF crime data. Figure \ref{fig:VAN-dynamics} displays the result of same experiment on VAN crime data. Similarly, we observe the general trend of moving to the bottom-right corner in the predictions across most event categories. This experiments confirms our conclusion that MAPSED captures the short-term dynamics information for prediction.

\subsection{Experiment Platform}
In our work, we conduct all of the experiments on a single workstation. The workstation configuration is shown in the following Table~\ref{tab:resources}.

For softwares, we use Python 3.8.8 and Pytorch 1.7.1.
\begin{table}[tb]
    \centering
    \caption{Summary of computational resource}
    \resizebox{\linewidth}{!}{
    \begin{tabular}{ccccc}
    \toprule
        Memory&Hard drive& CPU& GPU& Workstation \\\midrule
        64GB & 512GB SSD+2TB HDD&Intel Core i7-10700 & Nvidia RTX3090&Alienware Aurora R11 \\\bottomrule
    \end{tabular}}
    \label{tab:resources}
\end{table}

    